%% file: main.tex
\documentclass[letterpaper]{article} 
\usepackage{aaai2027}  
\usepackage[hyphens]{url}  
\usepackage{graphicx} 
\usepackage{natbib}  
\usepackage{caption} 
\usepackage{booktabs}
\usepackage{amsmath,amssymb}
\usepackage{multirow}
\usepackage{pifont}
\usepackage{cleveref}
\crefname{section}{Section}{Sections}
\Crefname{section}{Section}{Sections}
\crefname{figure}{Figure}{Figures}
\Crefname{figure}{Figure}{Figures}
\crefname{table}{Table}{Tables}
\Crefname{table}{Table}{Tables}
\usepackage{algorithm}
\usepackage{algorithmic}
\usepackage{newfloat}
\usepackage{listings}
\DeclareCaptionStyle{ruled}{labelfont=normalfont,labelsep=colon,strut=off} 
\floatstyle{ruled}
\newfloat{listing}{tb}{lst}{}
\floatname{listing}{Listing}

\newcommand{\finding}[1]{\paragraph{Finding.} #1}

\title{Same Task, Different Statement: When and How Reformulation Helps LLM Problem Solving}
\author{
    Wentao Zhang\textsuperscript{\rm 1,$\dagger$},
    Xiaoyu Zhang\textsuperscript{\rm 2,$\dagger$},
    Chengwei Liu\textsuperscript{\rm 3,*},
    Yutao Hu\textsuperscript{\rm 4},
    Tao Zhang\textsuperscript{\rm 1,*},
    Junsheng Wu\textsuperscript{\rm 1},
    Hongping Gan\textsuperscript{\rm 1},
    Yang Liu\textsuperscript{\rm 2}
}
\affiliations{
    \textsuperscript{\rm 1}Northwestern Polytechnical University\\
    \textsuperscript{\rm 2}CCDS, Nanyang Technological University\\
    \textsuperscript{\rm 3}Nankai University\\
    \textsuperscript{\rm 4}Huazhong University of Science and Technology
}
\copyrighttext{\textsuperscript{$\dagger$}These authors contributed equally to this work and share first authorship. \textsuperscript{*}Corresponding authors (tao\_zhang@nwpu.edu.cn; chengwei.liu@nankai.edu.cn).}

\begin{document}

\maketitle


\begin{abstract}
When people face a hard problem, they often restate it in a clearer way
before solving it. We study whether the same habit helps large language models by rewriting the statement of a problem. A legal rewrite must keep the meaning, the constraints, and the correct answer unchanged. We test three kinds of rewrites. Decomposition makes hidden subgoals explicit.
Typed Reformulation casts the task into a more suitable solving frame.
Concept Reduction removes surface detail and exposes the task structure.
We evaluate them on eight math, logic, and code benchmarks with eight
models, all under the same chain-of-thought solving prompt.
The effect is large but depends on the problem type. An oracle that picks the best rewrite after the fact raises accuracy by $+27.6$\% on average.
Yet no single rewrite helps every problem, and a mismatched rewrite can hurt.
Still, the effect follows clear patterns. Each kind of rewrite fits
certain problem types. A simple fixed rule learned from one set of
problems transfers part of the gain to new problems. It gains $+11.0$\%
on the problems it chooses to rewrite, while blind rewriting loses
$5.6$\% on a similar share of problems.
Rewriting is therefore not a universal way to solve problems. The question is when and how to rewrite.
\end{abstract}


\input{src/Intro}
\input{src/bg}
\input{src/framework}
\input{src/Experimental}
\input{src/Results}




\bibliography{aaai}

\end{document}

%% file: src/Intro.tex
\section{Introduction}
\label{sec:introduction}

\begin{figure}[!t]
\centering
\includegraphics[width=0.49\textwidth]{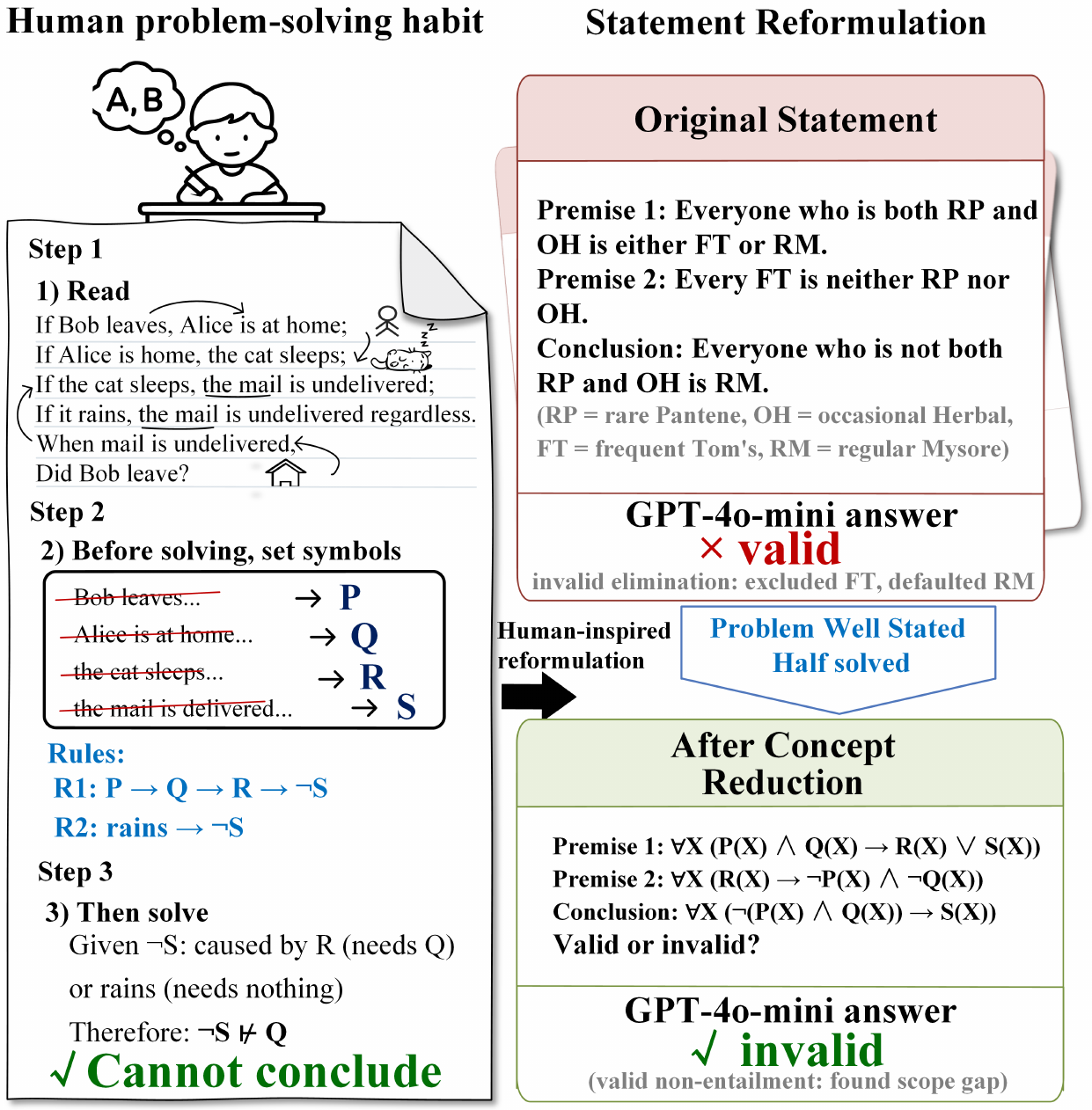}
\caption{A rewrite can unlock a solution the original statement misses.
Left: a human solver first rewrites a logic story as symbols and rules,
and only then solves it. Right: Concept Reduction applies the same habit
to an LLM input. The reformulated statement flips GPT-4o-mini from a wrong
\emph{valid} to a correct \emph{invalid}, while the task and its correct
answer are unchanged.}
\label{fig:teaser}
\end{figure}

Large language models (LLMs) are increasingly used to solve mathematical, logical, and programming problems \cite{cobbe2021training,hendrycks2021measuring}. Yet their performance remains unstable. A model may solve one problem correctly but fail on another that appears to require similar knowledge and reasoning. Such failures are usually attributed to limited reasoning ability or an ineffective solving strategy.

Human problem solving suggests another factor. People often do not work directly from the original statement of a difficult problem. They may break it into smaller parts, express it in a more explicit form, or remove surface details to reveal its underlying structure \cite{polya1945how,newell1958elements}. Although these operations leave the task unchanged, they can substantially affect its difficulty \cite{hayes1976understanding,kotovsky1985why,kintsch1988use}. Solving a problem therefore depends not only on carrying out the reasoning, but also on finding a representation that supports it.

Research on LLM problem solving has largely focused on the former. Chain-of-thought (CoT), self-refinement, search, and tool use improve how models generate, inspect, and execute solutions \cite{wei2022chain,madaan2023selfrefine,yao2023tree,schick2023toolformer}. The problem statement itself, however, is typically left unchanged. We know much less about what happens when the same task is decomposed, transformed, or abstracted into a different representation. Some apparent solver failures may therefore arise not from a lack of capability, but from a problem representation that the model cannot use effectively.

\Cref{fig:teaser} illustrates this possibility with two logic problems.
Left: a human solver first rewrites a story of implication rules as symbols. 
The symbolic view makes it clear that the backward question cannot be concluded.
Right: Concept Reduction applies the same habit to an LLM input.
Under the original wording, GPT-4o-mini wrongly answers \emph{valid} by
eliminating FT and defaulting to RM. After the statement is restated with explicit symbolic premises, the same
model finds the scope gap and correctly answers \emph{invalid}. The task and the correct answer remain unchanged. Only the
representation differs.

Motivated by this observation, we study when and how problem representation affects
LLM problem solving. We formalize reformulation as a legality-constrained transformation from a problem statement and solving goal $(x,g)$ to an alternative representation $(x',g')$, while withholding the evaluation target $a$. A legal reformulation must preserve the task meaning, constraints, goal, answer interface, and set of correct final answers (\cref{sec:framework}). We instantiate three reformulation families based on prior work in human problem solving \cite{polya1945how,newell1958elements,hayes1976understanding,kotovsky1985why,kintsch1988use}: \textbf{Decomposition (D)} reorganizes a problem into clearer components, \textbf{Typed Reformulation (TR)}  recasts the task under a more suitable solving frame, and \textbf{Concept Reduction (C)} removes incidental details to expose its underlying structure.

We evaluate these reformulations on 1,064 problems for analysis and another 1,064 held-out problems for transfer, drawn from eight math, logic, and code benchmarks. The problems cover nine types labeled by LLMs, with labels on the analysis set checked by humans. We test eight LLMs under the same zero-shot CoT prompt in all conditions, so the measured differences come from the problem representation rather than from a stronger solving strategy.

This study is organized around three research questions.

\textbf{RQ1: Effectiveness.}
Can a legal reformulation improve LLM accuracy on the same task, and by how much?
To study the impact of legal reformulation on the same task, we run
experiments across eight models. Accuracy on the original statements
ranges from 43.7\% to 63.6\%. Choosing the best legal reformulation per
problem after evaluation raises it to 80.0\% to 86.4\%. Matched controls
with extra Orig samples or free paraphrases close much less of this gap.

\textbf{RQ2: Preferences.}
Which reformulation works for which kind of problem?
To study when each reformulation helps, we compare each reformulation family
by source type. Decomposition helps Search and Classification; Concept Reduction
helps Explanation and Causality; Typed Reformulation helps Optimization,
Design, and Verification. A mismatch can hurt, and no single family fits all
types. This suggests that reformulation should be conditioned on the source type.

\textbf{RQ3: Applicability.}
Can these preferences guide reformulation selection on unseen problems?
To test whether these preferences transfer, we apply a simple selection rule on
held-out problems. The rule gains $+11.0$\% on the
problems it chooses to rewrite. Blind rewriting loses $5.6$\% at similar
coverage, and random rewriting stays near Orig ($+0.8$\%).

This paper makes four contributions.
\ding{182} We formalize problem reformulation as a legality-constrained transformation and instantiate three complementary families: Decomposition, Typed Reformulation, and Concept Reduction.
\ding{183} Through experiments on eight benchmarks, nine problem types, and eight LLMs, we show that equivalent representations can produce large accuracy differences and characterize when each family helps or hurts.
\ding{184} We turn these empirical preferences into a simple selection rule and show that part of the gain transfers to held-out problems.
\ding{185} We will release the original and reformulated problem variants, together with the artifacts needed to reproduce the main results.

%% file: src/bg.tex
\section{Preliminaries \& Related Work}
\label{sec:background}

\subsection{Human Problem Solving Methods}

Cognitive science treats problem solving as an active process of constructing a
usable representation from surface wording
\cite{polya1945how,newell1958elements}. Studies of problem isomorphs show that
semantically equivalent tasks can differ substantially in difficulty depending
on their presentation \cite{hayes1976understanding,kotovsky1985why}. For
example, a word problem can be harder than the same relations expressed in
equational form because the narrative adds comprehension load. Cognitive load
theory links this difficulty to maintaining many interacting relations at once.
When that load is high, performance drops
\cite{sweller1988cognitive,paas2003cognitive}. This motivates studying how
presentation changes LLM accuracy when the underlying task is fixed.

\subsection{LLM Problem Solving Methods}

Most work on improving LLM problem solving intervenes on the solver side.
Chain-of-thought prompting elicits longer reasoning traces
\cite{wei2022chain}. Self-refinement prompts the model to revise its own
outputs \cite{madaan2023selfrefine}. Search-based methods explore multiple
solution paths \cite{yao2023tree,besta2024graph}. Tool-augmented systems
delegate subtasks to external modules such as calculators, code executors, or
retrieval engines \cite{schick2023toolformer,parisi2022talm}. These methods
expand what LLMs can solve, but they mainly strengthen solution execution while
leaving the input statement fixed.

\begin{figure*}[t]
\centering
\includegraphics[width=0.95\textwidth]{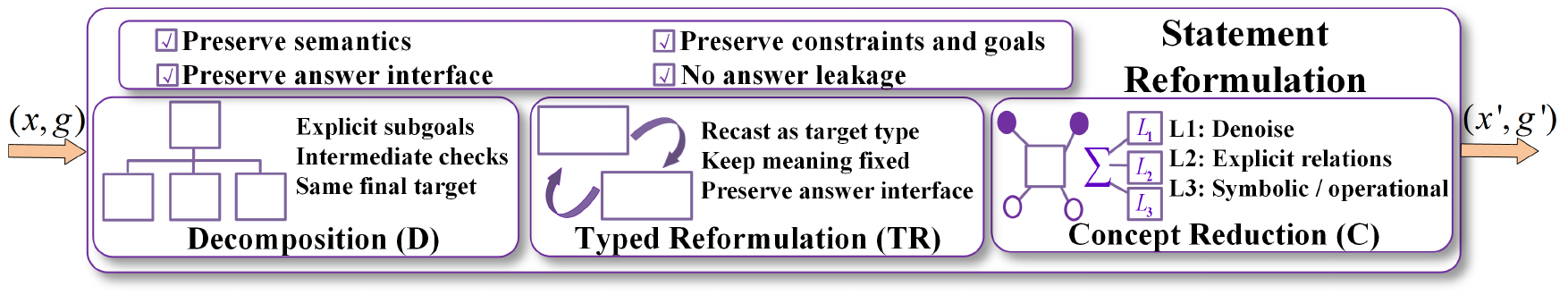}
\caption{Task-preserving statement reformulation. Rewrites map $(x,g)$ to
$(x',g')$ under shared legality constraints, using three controlled families:
Decomposition, Typed Reformulation, and Concept Reduction. The evaluation target $a$ is held out until evaluation.}
\label{fig:reformulation}
\end{figure*}

A smaller line of work also changes how the problem is stated before or while
solving. Some methods decompose the query into subquestions
\cite{zhou2023least}, step back to a higher-level principle or otherwise
abstract the stem \cite{zheng2024stepback}, or rephrase the question
\cite{deng2024rephrase}. These ideas are close in spirit to human
re-representation. In practice, they are usually one-shot heuristics inside the
solver prompt. However, they typically do not combine three controls that we
require here: isolating presentation change from solving, enforcing explicit
task-preservation constraints across controlled reformulation families, and
measuring when a rewrite helps or hurts by problem type. We reformulate the
statement itself, hold the downstream CoT prompt fixed, and evaluate
legality-constrained families at scale.

%% file: src/framework.tex
\section{Statement Reformulation}
\label{sec:framework}

We study statement-level reformulations that change how a problem is presented
without changing the underlying task. We distinguish the problem presentation
from the evaluation target. The presentation consists of the natural-language
statement $x$ and the solving goal or instruction $g$. A labeled evaluation item
additionally includes an evaluation target $a$, such as a final answer or a
functional test suite, which is used only for evaluation.

A reformulation receives only the presentation $(x,g)$ and maps it to a
rewritten presentation
$
\Phi(x,g)=(x',g')
$.
At evaluation time, the rewritten presentation $(x',g')$ is paired with the
same evaluation target $a$ as the original item. Thus, the reformulator does
not have access to the answer or test suite.
To separate presentation effects from task changes, we first define when a
reformulation is legal. We then instantiate three human-inspired reformulation
families under these
criteria, as summarized in \Cref{fig:reformulation}.

\subsection{Task-Preservation Constraints}
\label{subsec:legality}

To measure presentation effects rather than task changes, we require every
reformulation to preserve the original task. Throughout this paper, we call a reformulation legal if it satisfies the following
conditions:

\ding{182} \textbf{Preserve semantics.} The rewritten presentation preserves the entities, quantities, units, relations,
and assumptions needed to interpret the task.
\ding{183} \textbf{Preserve constraints and goals.} It does not drop, weaken,
    add, or replace operative constraints, and it asks for the same objective
    as the original problem.
\ding{184}\textbf{Preserve answer interface.} The rewrite preserves the set of
    correct final answers and the required answer type, precision, and expected
    format.
\ding{185}\textbf{No answer leakage.} The rewrite is constructed from $(x,g)$
    alone. It must not use the evaluation target $a$ or insert oracle intermediate
    results that would require solving the problem.

These constraints do not require the rewrite to keep the same solving path.
Decomposition and Typed Reformulation may add subgoals or change the stated
solving frame, as long as the task and answer interface stay the same. 

\subsection{Decomposition}
To test whether hidden dependency chains contribute to presentation-sensitive
failures, Decomposition makes intermediate objectives explicit. The idea is
motivated by subgoaling, a standard strategy in human problem solving
\cite{polya1945how,newell1958elements}. It is intended to expose dependencies
that are already implicit in the original statement, rather than to add new information or change the task.
For a problem presentation $(x,g)$, we construct a subgoal scaffold
\[
\mathcal{S}_D(x,g)=\{(x_i,g_i)\}_{i=1}^n,
\]
where each $(x_i,g_i)$ specifies an intermediate check or objective. The
scaffold does not provide the answers to these subgoals. We then serialize the
scaffold together with the original task into a single rewritten presentation
\[
\Phi_D(x,g)=(x'_D,g'_D).
\]
A legal Decomposition reformulation may make the order and dependencies among
subgoals explicit, but it must not add assumptions, constraints, or facts that
are not entailed by the original statement.

\subsection{Typed Reformulation}

To test whether the choice of reasoning frame affects LLM problem solving,
Typed Reformulation recasts the same task as a different type of reasoning.
The idea is motivated by evidence that problem formulation can affect
difficulty even when the underlying task is preserved
\cite{hayes1976understanding,kotovsky1985why}. Admissible frames may include
computation, verification, planning, or optimization.

For a problem presentation $(x,g)$ and an admissible reasoning frame $r_j$,
we define a typed reformulation as
\[
\Phi_{TR}^{(j)}(x,g)=(x'_{TR,j},g'_{TR,j}).
\]
The reformulation may change which structure is made salient, but it must preserve the original semantics,
constraints, objective, answer type, and output format. It must not introduce a new assumption, relaxation, or solution.

\begin{figure}[t]
\centering
\includegraphics[width=0.49\textwidth]{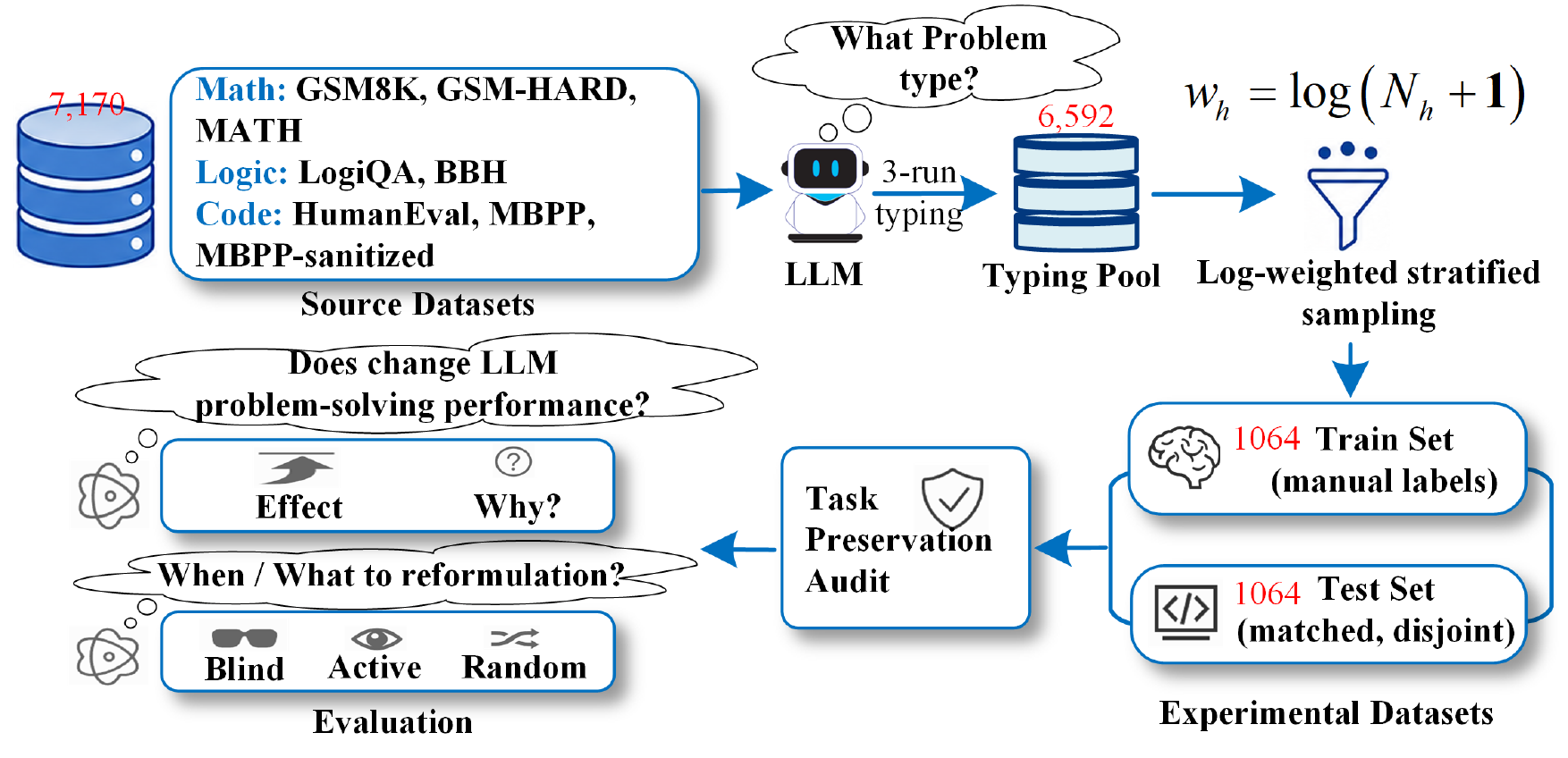}
\caption{Study design. From eight public benchmarks we form a typed pool and
sample a Train Set and a disjoint Test Set. Candidate reformulations
pass a task-preservation audit. Evaluation uses eight LLMs with a shared zero-shot CoT prompt.}
\label{fig:study_design}
\end{figure}

\subsection{Concept Reduction}

To test whether surface detail can obscure the relations needed for solving,
Concept Reduction removes incidental presentation detail or makes
task-relevant structure more explicit. The idea is motivated by theories of
comprehension and abstraction
\cite{kintsch1988use,bruner1960process}. We define
\[
\Phi_C^{(k)}(x,g)=(x'_C{}^{(k)},g'_C{}^{(k)}),
\qquad k \in \{1,2,3\},
\]
where $k$ indexes increasing levels of abstraction.

At level L1 ($k=1$), the reformulation removes non-essential narrative while
retaining the original facts. At level L2 ($k=2$), it makes the key entities,
quantities, and relations explicit. At level L3 ($k=3$), it moves toward a
symbolic or operational presentation. These levels may change how the task is
represented, but they must not remove or alter any fact, relation, assumption,
constraint, or answer requirement needed to solve the original task. 

\subsection{Evaluation Quantities}
\label{subsec:eval_quantities}

Let $M$ be a fixed LLM solver (shared zero-shot CoT protocol as in
\cref{sec:setup}). For any presentation, $\mathrm{Acc}_M$ is $1$ if $M$
answers correctly under the held-out evaluation target $a$, and $0$
otherwise. \emph{Orig} is the mean of $\mathrm{Acc}_M$ on the original
presentations.

For each family $F\in\{\mathrm{C},\mathrm{D},\mathrm{TR}\}$, let
$\mathcal{V}_F(x,g)$ be its legal rewrites of $(x,g)$ (one Decomposition; one
or more Typed Reformulation paths; Concept Reduction at levels $k{=}1,2,3$).
We report three nested retrospective ceilings per problem: Family Best
(any rewrite in $F$), Family Oracle (Best of $F$, or keep Orig), and Oracle
Selection (``Sel.''; Orig or Best across families). Formally,
\begin{align*}
\mathrm{Best}_F(x,g)
&= \max_{(x',g')\in\mathcal{V}_F(x,g)} \mathrm{Acc}_M(x',g'), \\
\mathrm{Oracle}_F(x,g)
&= \max\bigl\{\mathrm{Acc}_M(x,g),\,\mathrm{Best}_F(x,g)\bigr\}, \\
\mathrm{Sel}(x,g)
&= \max\bigl\{\mathrm{Acc}_M(x,g),\,
\max_F \mathrm{Best}_F(x,g)\bigr\}.
\end{align*}
Averaging Family Best over problems, then taking the best of the three family
means, gives \emph{MS-Best}. Averaging $\mathrm{Sel}$ gives the Oracle
Selection column.




%% file: src/Experimental.tex
\section{Study Setup}
\label{sec:setup}

\Cref{fig:study_design} summarizes the overall study design.

\subsection{Datasets and Sampling}

The eight source benchmarks cover three domains. Math problems come from GSM8K
\cite{cobbe2021training}, GSM-HARD \cite{gao2022pal}, and MATH500
\cite{hendrycks2021measuring}. Logic problems come from BBH
\cite{suzgun2022challenging} and LogiQA \cite{liu2020logiqa}. Code problems
come from HumanEval \cite{chen2021evaluating}, MBPP-full
\cite{austin2021program}, and MBPP-sanitized \cite{austin2021program}.
We define 12 initial source types, adapted from classic problem-solving
classifications \cite{newell1958elements}. The LLM judge (GPT-4o) assigns each problem a type independently in each of three runs. Three types are
merged or excluded because of small size or low inter-run consistency,
leaving the nine types used below. The
resulting typing pool contains 6,592 problems, which is 91.9\% of the original
7,170.

The pool is dominated by Computation. We therefore use disproportionate
stratified sampling over domain and source-type pairs so that rare types retain
adequate coverage for conditional analysis. For each pair $h$, let $N_h$ be the
number of problems in the typing pool. We set stratum weights
$w_h=\log(N_h+1)$ and take target sample counts proportional to $w_h$, subject to
availability. This oversamples rare types relative to proportional allocation
while still reflecting pool size. The total size of each set (1,064) is chosen
so that the 95\% Wilson interval half-width for a single proportion is
approximately $3.0$\% in the most conservative case of
$p=0.5$~\cite{wilson1927probable}.
\Cref{tab:source_type_distribution} reports the resulting counts.

\begin{table}[t]
\centering
\small
\setlength{\tabcolsep}{6pt}
\begin{tabular}{l|r|r|r}
\hline
\textbf{Source Type} & \textbf{Typing Pool} & \textbf{Train Set} & \textbf{Held-out Test} \\
\hline
Search & 489 & 129 & 144 \\
Classification & 411 & 125 & 138 \\
Computation & 3,980 & 167 & 197 \\
Explanation & 498 & 129 & 144 \\
Causality & 190 & 110 & 80 \\
Optimization & 205 & 112 & 93 \\
Planning & 46 & 46 & 0 \\
Design & 286 & 118 & 125 \\
Verification & 487 & 128 & 143 \\
\hline
\textbf{Total} & 6,592 & 1,064 & 1,064 \\
\hline
\end{tabular}
\caption{Source-type distribution after 3-run consistency classification and
log-weighted stratified sampling. \scriptsize{Types are operational labels for the stem's
main demand: Search (locate/filter), Classification (choose among options),
Computation (calculate), Explanation (account for a fact), Causality (judge
cause), Optimization (best choice under constraints), Planning (schedule under
constraints), Design (construct an artifact), Verification (check correctness).}}
\label{tab:source_type_distribution}
\end{table}

The Train Set is used for the main RQ1 and RQ2 analyses. Its \emph{source-type}
labels are manually validated: auditors reviewed all 1,064 problems and
corrected 39 label errors (3.7\%).
The held-out Test Set contains another 1,064 problems drawn from the remaining
pool by the same log-weighted rule. The two sets are strictly disjoint.
Planning has no Test Set items (\cref{tab:source_type_distribution}), so
Planning-specific claims are Train-only. RQ3 uses the Test Set without access
to its source-type labels (\cref{subsec:rq3_probes}).

\subsection{Reformulation Generation}

For each problem presentation $(x,g)$, an LLM produces candidates from the three
families in \cref{sec:framework}, using $(x,g)$ only. The evaluation target $a$
is held out.

\subsection{Task-Preservation Audit}

Each candidate then passes a three-stage audit before entering the
validated-only track used in the main tables.

\textbf{Rule-based screening}
checks format violations and obvious preservation failures such as missing
numbers, changed units, or inserted final answers.
\textbf{Independent LLM review} uses GPT-4o \cite{openai2024gpt4o} as a judge (separate from the eight
solvers) to compare the original and rewritten presentations and to flag
semantic drift, dropped or added constraints, answer-interface changes, or
answer leakage. Failed items are regenerated within a fixed retry budget.
\textbf{Human audit} applies stratified single-annotator checks across
datasets, families, and Concept Reduction levels. On 240 validated Train Set
candidates, two annotators independently labeled each rewrite as legal or
illegal (\cref{tab:audit_agreement}). Raw agreement is the share of items with
the same label. Cohen's $\kappa$ adjusts that rate for chance agreement, so
$\kappa{=}1$ is perfect agreement beyond chance and $\kappa{=}0$ is no better
than chance. Overall they reach 95.0\% agreement ($\kappa{=}0.72$). Agreement
is perfect on D and high on TR; on C, raw agreement stays high (91.2\%) but
$\kappa$ drops because illegal labels are rare, which inflates chance
agreement. Test Set reformulations follow the same audit protocol.

\begin{table}[t]
\centering
\small
\setlength{\tabcolsep}{18pt}
\begin{tabular}{lrrr}
\hline
\textbf{Slice} & \textbf{$n$} & \textbf{Agreement} & \textbf{$\kappa$} \\
\hline
All & 240 & 95.0\% & 0.72 \\
C & 80 & 91.2\% & 0.34 \\
D & 80 & 100.0\% & 1.00 \\
TR & 80 & 93.8\% & 0.82 \\
\hline
\end{tabular}
\caption{Dual-annotator legality labels on 240 Train Set candidates.}
\label{tab:audit_agreement}
\end{table}

\subsection{Models and Inference}

We evaluate eight language models: GPT-4o-mini and GPT-5-mini
\cite{openai2024gpt4o,openai2025gpt5}, Qwen3-Flash and Qwen3-Plus
\cite{qwen2025qwen3}, Llama-3.1-8B and Llama-3.1-70B
\cite{meta2024llama3}, and DeepSeek-V4-Flash and DeepSeek-V4-Pro from
DeepSeek-AI (API models distinct from DeepSeek-R1;
\cite{guo2025deepseekr1}). All models are queried with greedy decoding
(temperature $=0$, top-p $=1$) under a unified zero-shot CoT solving prompt
shared by Orig and all reformulated variants. Max tokens are set by dataset.
Each problem is evaluated once per model and variant for the main results.

\subsection{Evaluation Metrics}

The primary metric is accuracy. For math and logic, we compare the extracted
final answer to the ground truth under the evaluator's matching rules. For code
tasks (HumanEval / MBPP), correctness is functional. The generated function is
executed against the benchmark unit tests under the original signature
(pass@$1$: one greedy sample per problem \cite{chen2021evaluating}). Syntax
errors, runtime errors, and timeouts count as incorrect.

For paired comparisons between original and reformulated presentations, we use
McNemar's exact test \cite{mcnemar1947note}. We report these tests for the main
RQ1 Best/Oracle contrasts, the RQ2 type$\times$family contrasts, and the RQ3 selection comparisons. We also report accuracy deltas and
win/loss counts. Unless otherwise stated, the main tables report the
validated-only track.

\subsection{Settings for Applicability Test}
\label{subsec:rq3_probes}

With the Train Set, Test Set, reformulations, and solvers fixed as above, RQ3
compares three frozen selection settings on the Test Set. All three see only the
problem text and domain. None sees
Test Set source-type labels, ground-truth answers, or whether a candidate
reformulation would be answered correctly. Decisions are made once and then
shared across the eight solvers. Preference tables and filter rules are frozen
from the Train Set before any Test Set evaluation.

\textbf{Blind.} An LLM (GPT-4o) chooses Orig, C (with level), D, or TR from the
problem text and domain alone. It knows the three families, but it receives no
Train Set preference table and no outcome feedback. 

\textbf{Random.} For each Test Set problem, a fixed-seed random draw matches
Active's overall coverage and C/D/TR mix, then applies the same filter. This
isolates the filter from Train Set preferences.

\textbf{Active.} The same LLM interface as Blind, but the prompt also includes a
frozen Train Set preference pack: the type-level Rec marked by $\star$ in
\cref{fig:rq2_delta} (construction in \cref{sec:rq2}), plus short domain
biases. At runtime Active may infer a source type to look up that prior; that
inference can be wrong, in which case the looked-up Rec may not match the stem.
When confidence is low, the stem is already short and explicit, or the preferred
family is a poor fit, Active defaults to Orig. After a non-Orig choice, a
legality and quality filter checks the converted statement. Failures fall back
to Orig.

%% file: src/Results.tex
\section{Results \& Analysis}
\label{sec:results}

\begin{table*}[t]
\centering
\scriptsize
\setlength{\tabcolsep}{9pt}
\begin{tabular}{l|c|cc|cc|cc|cc|cc}
\hline
& & \multicolumn{2}{c|}{\textbf{Concept Reduction}} & \multicolumn{2}{c|}{\textbf{Decomposition}} & \multicolumn{2}{c|}{\textbf{Typed Reformulation}} & \multicolumn{2}{c|}{\textbf{MS-Best}} & \multicolumn{2}{c}{\textbf{Oracle Selection}} \\
\textbf{Model} & \textbf{Orig} & \textbf{C-Best} & \textbf{C-Oracle} & \textbf{D-Best} & \textbf{D-Oracle} & \textbf{TR-Best} & \textbf{TR-Oracle} & \textbf{Acc} & \textbf{$\Delta$} & \textbf{Sel.} & \textbf{$\Delta$} \\
\hline
GPT-4o-mini & 43.7 & 55.1 & 58.0 & 63.0 & 67.3 & 58.5 & 60.0 & 63.0 & +19.3 & 80.7 & +37.0 \\
GPT-5-mini & 63.4 & 72.9 & 77.9 & 64.4 & 73.1 & 67.1 & 80.5 & 72.9 & +9.5 & 86.4 & +23.0 \\
Qwen3-Flash & 58.4 & 72.8 & 76.4 & 63.2 & 69.0 & 60.5 & 76.6 & 72.8 & +14.4 & 85.4 & +27.0 \\
Qwen3-Plus & 62.8 & 72.7 & 76.3 & 64.0 & 70.6 & 63.6 & 79.0 & 72.7 & +9.9 & 84.4 & +21.6 \\
DeepSeek-V4-Flash & 59.5 & 71.0 & 75.4 & 60.8 & 68.3 & 61.5 & 76.9 & 71.0 & +11.5 & 84.6 & +25.1 \\
DeepSeek-V4-Pro & 63.6 & 72.3 & 77.1 & 60.9 & 70.7 & 70.2 & 80.4 & 72.3 & +8.7 & 84.9 & +21.3 \\
Llama-3.1-8B & 42.5 & 62.4 & 68.3 & 48.9 & 59.6 & 61.1 & 68.6 & 62.4 & +19.9 & 80.0 & +37.5 \\
Llama-3.1-70B & 53.7 & 69.4 & 74.3 & 58.8 & 67.2 & 64.5 & 65.9 & 69.4 & +15.7 & 82.1 & +28.4 \\
\hline
\textbf{Mean} & \textbf{56.0} & \textbf{68.6} & \textbf{73.0} & \textbf{60.5} & \textbf{68.2} & \textbf{63.4} & \textbf{73.5} & \textbf{69.6} & \textbf{+13.6} & \textbf{83.6} & \textbf{+27.6} \\
\hline
\end{tabular}
\caption{RQ1 main results on the Train Set.
\scriptsize{(Metrics follow \cref{subsec:eval_quantities}. Sel.\ is Oracle
Selection. Both $\Delta$ columns are vs.\ Orig. Best and Oracle Selection are
retrospective ceilings.)}}
\label{tab:rq1_main}
\end{table*}

\subsection{RQ1: Effectiveness}
\label{sec:rq1}

To evaluate whether a legal rewrite can raise LLM accuracy on the same task,
we compare Orig with $\mathrm{Best}_F$, $\mathrm{Oracle}_F$, MS-Best, and
Oracle Selection on the Train Set across eight solvers
(\cref{subsec:eval_quantities,tab:rq1_main}).

\medskip\noindent\textbf{Analysis.}
Within-family Best accuracies are already higher than Orig: mean C/D/TR-Best
are 68.6\%, 60.5\%, and 63.4\% against a 56.0\% Orig mean. Exact McNemar tests
find C-Best above Orig for all eight models ($p{<}0.001$); D-Best and TR-Best
for 5/8 ($p{<}0.05$). Which family peaks is solver-dependent (e.g., GPT-4o-mini
at D-Best 63.0\%; most others at C-Best). MS-Best averages 69.6\%
($+13.6$\% over Orig).
Family Oracle, which may keep the original statement, raises means to
73.0\% / 68.2\% / 73.5\%. The Best-to-Oracle gap means a forced rewrite often
flips an Orig-correct case, so the safe choice is sometimes to keep Orig. Cross-family Oracle Selection then reaches 83.6\%
($+27.6$\%; McNemar $p{<}0.001$ for all eight models), an upper bound if the
right family or variant can be chosen per problem.

To check that this headroom is not just extra attempts or unconstrained
rewording, we run matched-budget controls on GPT-4o-mini
(\cref{tab:rq1_passk}), outside the main once-per-variant protocol.
pass@$3$/pass@$6$ reach only 50.3\% / 54.8\%, and the
paraphrase oracle 60.3\%, versus 76.0\% under the family ceiling. Extra Orig
tries and free rewording therefore help much less than choosing among the
stored C/D/TR rewrites. Free paraphrases alone also underperform Orig.
Oracle Selection on this model is 80.7\%. It can exceed the family ceiling when
more Typed Reformulation candidates are available than the fixed stored set
used for that ceiling.

\begin{table}[t]
\centering
\small
\setlength{\tabcolsep}{25pt}
\begin{tabular}{l|c}
\hline
\textbf{Setting} & \textbf{Acc (\%)} \\
\hline
Orig & 43.7 \\
pass@$3$ & 50.3 \\
pass@$6$ & 54.8 \\
Paraphrase oracle & 60.3 \\
Family ceiling & 76.0 \\
Oracle Selection (\cref{tab:rq1_main}) & 80.7 \\
\hline
\end{tabular}
\caption{RQ1 matched-budget controls on GPT-4o-mini.
\scriptsize{(Under pass@$k$, an item is correct if any of $k$ independent Orig solves is
correct. The paraphrase oracle uses the original statement and five free
paraphrases, while the family ceiling uses the original statement and all stored C/D/TR reformulations.)}}

\label{tab:rq1_passk}
\end{table}

\begin{figure}[t]
\centering
\includegraphics[width=\linewidth]{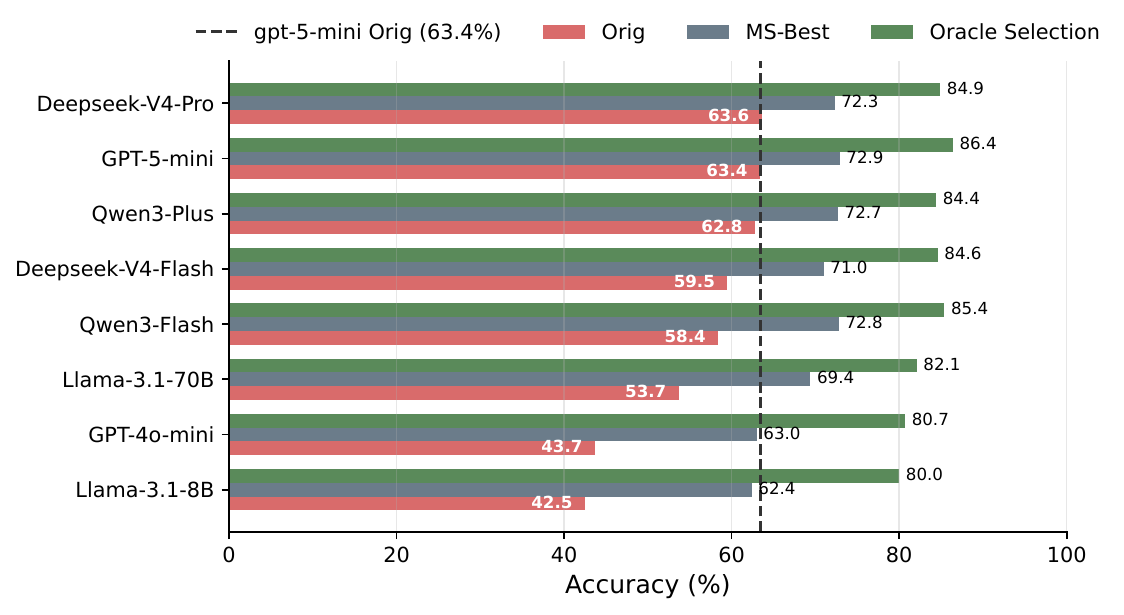}
\caption{RQ1 accuracy (\%) on the Train Set under Orig, MS-Best, and Oracle
Selection. Weaker models, including open ones such as Llama-3.1-8B, already
approach or pass the GPT-5-mini Orig baseline under
MS-Best. Oracle Selection lifts every model well above that line, so the
presentation ceiling far exceeds a strong closed model's original accuracy.
\scriptsize{The dashed line marks GPT-5-mini Orig (63.4\%) as a reference.}}
\label{fig:rq1_gap}
\end{figure}

\Cref{fig:rq1_gap} shows the practical implication: with Oracle Selection,
every model exceeds the GPT-5-mini Orig baseline, and several do so already
under MS-Best. Relative gains are larger for weaker solvers and smaller for
stronger ones, yet strong models still keep sizable Oracle Selection
headroom. Weaker models appear to rely more on statement scaffolding; stronger
models still leave presentation-sensitive failures on the table.

\finding{Some failures on the original statement look sensitive to presentation rather than pure capability gaps. Under a fixed CoT, changing only the statement can raise or lower accuracy. The size of the change depends on
family and model. Matched multi-sampling and free paraphrasing do not close the structured-family gap.}

\subsection{RQ2: Preferences}
\label{sec:rq2}

\begin{figure}[t]
\centering
\includegraphics[width=0.82\linewidth]{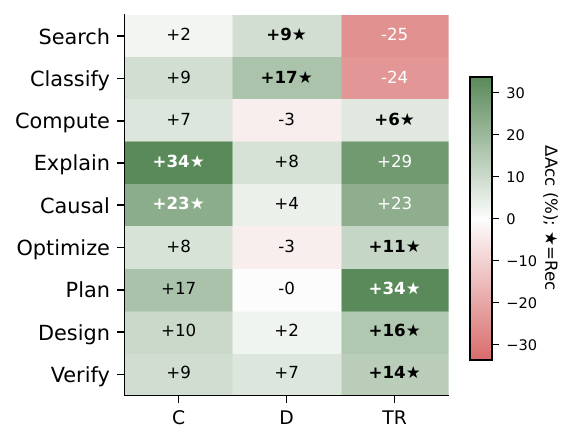}
\caption{RQ2 pooled $\Delta\mathrm{Acc}$ (\%) by source type and family.
\scriptsize{(Rows are source types, and columns are families. $\star$ marks Rec (\cref{sec:rq2}). Red-toned cells indicate harm.)}}

\label{fig:rq2_delta}
\end{figure}

To evaluate which rewrite fits which kind of problem, we compare
$\Delta\mathrm{Acc}(F)=\mathrm{Best}_F(x,g)-\mathrm{Acc}_M(x,g)$ by source type
on the Train Set, pooled over eight models
(\cref{subsec:eval_quantities,fig:rq2_delta}).

\medskip\noindent\textbf{Analysis.}
Three patterns stand out. Decomposition is most effective for structural
tasks: Search and Classification gain $+8.7$\% and $+16.7$\%, while Typed
Reformulation strongly hurts both ($-24.6$\% and $-23.6$\%). Recasting those
problems into another solving frame often breaks the dependency structure the
model needs, so TR is a mismatch rather than a mild rewrite. Concept Reduction
is strongest for abstraction-heavy tasks: Explanation and Causality gain
$+33.7$\% and $+23.2$\%, where stripping narrative exposes the operative
skeleton. Typed Reformulation is strongest for Optimization, Design, and
Verification; Planning also favors TR on the Train Set ($+33.6$\%), where a
better goal frame matches the task. Computation is mixed ($+6.8$\% for C vs.\
$+6.0$\% for TR in the pool). Pooled over models, exact McNemar tests support
these preferred matches (D on Search/Classification; C on Explanation/Causality;
TR on Optimization, Design, and Verification; all $p{<}0.001$) and the
Search/Classification harms under TR ($p{<}0.001$).

Two GPT-4o-mini Train Set cases show the same rewrite family helping or hurting.

\textbf{Help.}
A coding task asks to add a tuple to a list. Typed Reformulation turns the
original command into a question about what the list should look like
afterward. Under the original wording, the model changes the list in
place with \texttt{extend} and fails the test. After the rewrite, it
returns a new list and passes.

\textbf{Hurt.}
A workplace question asks how a typical person would judge the cause.
Typed Reformulation turns it into a policy-style evaluation with extra
scenario detail. The model answers the original wording correctly. After
the rewrite, it misses the counterfactual cue and fails.

The new statement helps when it fits the problem and hurts when it does not. Because effects depend on problem type, we turn the pooled Train Set pattern
into a recommended family (Rec) for RQ3. A Win is an original miss that Family
Best fixes; a Loss is the reverse. Within each domain we take the family with
largest $\mathrm{Net}=\mathrm{Wins}-\mathrm{Losses}$, then an $n$-weighted
majority over domains. Rec is frozen before RQ3. For Computation, the
domain-weighted Rec still selects TR because math and code dominate the pool
and both lean toward a clearer procedural frame.

\finding{Reformulation is not universally beneficial. The same family can help one source type and hurt another. Structural problems gain from making subgoals explicit, narrative problems from exposing the skeleton, and goal-framed problems from a better solving frame.  No single family fits all types. Which rewrite helps depends on the problem type.}

\subsection{RQ3: Applicability}
\label{sec:rq3}

To evaluate whether Train Set preferences transfer as a usable selection
method, we compare Orig, Blind, Random, and Active on the held-out Test Set
(\cref{fig:study_design,subsec:rq3_probes,tab:rq3_main}).

\begin{table}[!ht]
\centering
\scriptsize
\setlength{\tabcolsep}{2.5pt}
\begin{tabular}{l|cccc|ccc}
\hline
& \multicolumn{4}{c|}{\textbf{Full-set Acc (\%)}} &
\multicolumn{3}{c}{\textbf{$\Delta$Acc on acted-upon}} \\
\textbf{Model} & \textbf{Orig} & \textbf{Blind} & \textbf{Random} & \textbf{Active} &
\textbf{Blind} & \textbf{Random} & \textbf{Active} \\
\hline
GPT-4o-mini & 58.5 & 58.6 & 61.2 & 60.2 & $-$2.4 & +10.9 & +9.5 \\
GPT-5-mini & 68.8 & 65.8 & 67.7 & 68.9 & $-$15.4 & +0.5 & +2.0 \\
Qwen3-Flash & 64.9 & 63.3 & 63.8 & 66.3 & $-$10.6 & +1.0 & +7.0 \\
Qwen3-Plus & 69.0 & 69.4 & 68.3 & 70.5 & $-$0.5 & $-$2.1 & +8.0 \\
DeepSeek-V4-Flash & 66.7 & 65.9 & 62.5 & 68.0 & $-$1.9 & $-$0.5 & +6.5 \\
DeepSeek-V4-Pro & 66.9 & 64.9 & 67.1 & 68.1 & $-$11.5 & $-$2.1 & +6.5 \\
Llama-3.1-8B & 41.6 & 34.9 & 42.2 & 47.5 & $-$1.0 & $-$2.6 & +31.2 \\
Llama-3.1-70B & 57.5 & 57.1 & 58.1 & 60.7 & $-$1.4 & +1.0 & +17.1 \\
\hline
\textbf{Mean} & \textbf{61.7} & \textbf{60.0} & \textbf{61.4} & \textbf{63.8} &
\textbf{$-$5.6} & \textbf{+0.8} & \textbf{+11.0} \\
\hline
\end{tabular}
\caption{RQ3 on the Test Set. \scriptsize{(Left block: full-set accuracy (\%).
Right block: Acc(method)${-}$Acc(Orig) only on problems that method reformulates
(acted-upon; Blind $n{=}208$, 19.5\%; Random $n{=}193$; Active $n{=}199$, 18.7\%.)}}
\label{tab:rq3_main}
\end{table}

\medskip\noindent\textbf{Analysis.}
Active improves over Orig for seven of eight models (mean +2.0\%) and over
Blind for all eight (mean +3.8\%). Exact McNemar tests find Active higher than
Orig for 7/8 models and higher than Blind for 5/8 ($p{<}0.05$).
The full-set gain is modest because Active rewrites only 18.7\% of Test
problems and otherwise keeps Orig; on the acted-upon subset (the problems each
method chooses to rewrite) the conditional gain is large (+11.0\%), while Blind
hurts ($-$5.6\%) at similar coverage. On
Orig-correct problems, the correct-to-incorrect flip rate is 1.3\% for Active
versus 13.0\% for Blind, so Blind's damage is mostly harmful mismatches.
Gains are largest in logic (+3.0\%), where RQ2 favors C/D-style clarification;
smaller in math ($+0.5$\%); and near zero in code, where many stems are already
short and Active more often keeps the original statement. Weaker models show
larger acted-upon lifts (e.g., Llama-3.1-8B), while strong models move less but
still avoid Blind's high flip rate. Random stays near Orig on the full set
(61.4 vs.\ 61.7) and gains only +0.8\% when it rewrites.

\finding{Train Set preferences can be turned into a selection method that
recovers part of the presentation effect on held-out problems. Blind rewriting
at similar coverage hurts the problems it acts on. Active helps them and flips
far fewer Orig-correct cases. Random does not close this gap.}

\section{Limitations and Future Work}
\label{sec:discussion}

Our study has three limits of scope:
\ding{182} \textbf{Single-family interventions.}
We test one kind of reformulation at a time. This keeps the effect of each kind
separate and clear. Combining different kinds, or applying them one after
another, may bring extra gains or harms that our design cannot see. We
leave such combinations to future work.
\ding{183} \textbf{Testing within the same mixture.}
Our Test Set problems come from the same eight benchmarks as the Train Set
problems. The results therefore show that preferences learned on the Train Set
carry to new problems within this mixture. They do not show whether the
preferences also carry to a new domain.
\ding{184} \textbf{Simple selection rule.}
Active is a transfer probe built from Train Set preferences. It may infer the
wrong source type when looking up Rec. Stronger selection methods, and
combinations with tools or search, are natural next steps.

\section{Conclusion}
\label{sec:conclusion}

We studied statement-side reformulation for LLM problem solving, using
three reformulation families drawn from human problem solving.  Under a shared chain-of-thought solving prompt, how a problem is stated can change accuracy by a large margin even when the task itself does not change. The effect is conditional on problem type, domain, and model.  The broader implication is that some apparent capability gaps are presentation gaps.
For LLMs as for people, a problem well stated is often half solved.  We release the original and reformulated variants so that others can study when and how to restate.